\documentclass[lettersize,journal]{IEEEtran}
\usepackage{amsmath,amsfonts}
\usepackage{algorithmic}
\usepackage{algorithm}
\usepackage{array}
\usepackage[caption=false,font=normalsize,labelfont=sf,textfont=sf]{subfig}
\usepackage{textcomp}
\usepackage{stfloats}
\usepackage{url}
\usepackage{verbatim}
\usepackage{graphicx}
\usepackage{cite}
\usepackage{wrapfig}
\usepackage{colortbl}
\usepackage{booktabs}
\usepackage{xcolor}
\definecolor{blue}{rgb}{0,0,1}
\definecolor{mycolor}{HTML}{E0EBFF}

\usepackage{amsmath}
\usepackage{amssymb}
\usepackage{booktabs}
\usepackage{multirow}
\usepackage{makecell}
\usepackage{balance} 
\usepackage{soul}
\usepackage[accsupp]{axessibility}
\usepackage{hyperref}
\hypersetup{
    colorlinks=true,
    linkcolor=blue,
    anchorcolor=blue,
    citecolor=blue
}

\begin{document}

\title{GT23D-Bench: A Comprehensive General Text-to-3D Generation Benchmark}

\author{
        Xiao Cai$^{1}$,
        Sitong Su$^{1}$,
	Jingkuan~Song$^{2}$,
        Pengpeng Zeng$^{2}$,
        Ji Zhang$^{3}$,
        Qinhong Du$^{1}$,
        Mengqi Li$^{1}$,
 Heng~Tao~Shen$^{1,2}$, and Lianli~Gao$^{1}$\\
        
        $^{1}$University of Electronic Science and Technology of China \\
        $^{2}$Tongji University \quad \quad
        $^{3}$Southwest Jiaotong University



}

\markboth{}%
{Shell \MakeLowercase{\textit{et al.}}: Bare Demo of IEEEtran.cls for Computer Society Journals}



\maketitle

\begin{figure*}
    \centering
    \includegraphics[width=0.95\linewidth]{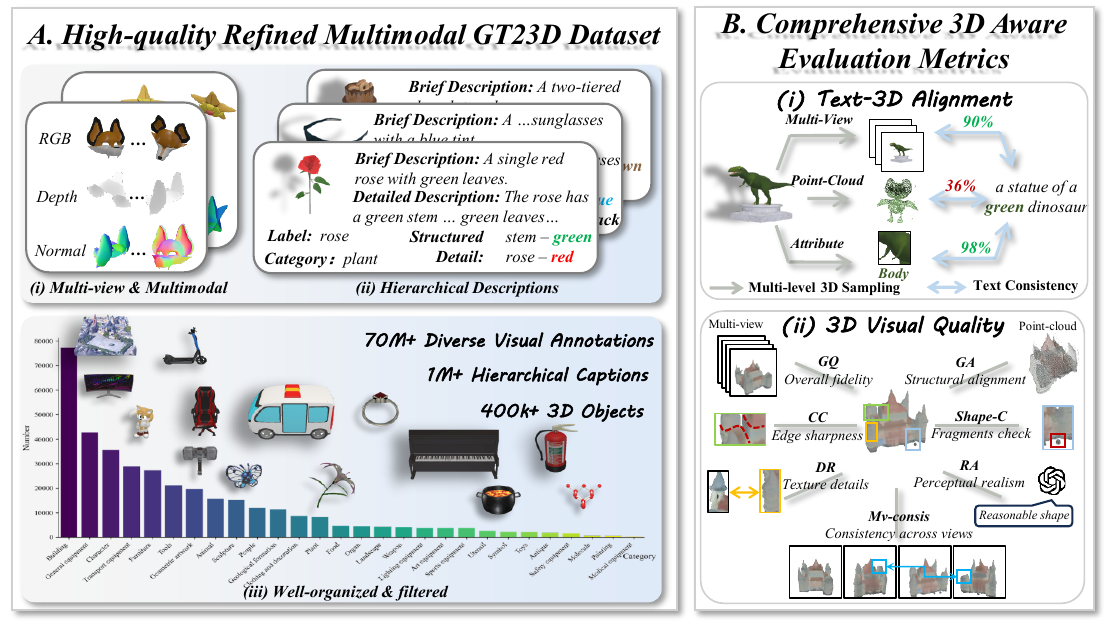}
    \caption{\textbf{Illustration of GT23D-Bench.} GT23D-Bench is the first benchmark for General Text-to-3D, which consists of two components: 1) a 400K multimodal-annotated label-organized thoroughly-filtered 3D Dataset (left) and 2) comprehensive 3D-Aware Evaluation Metrics (right).}
    \label{fig:frontpage}
\end{figure*}

\begin{abstract}

Text-to-3D (T23D) generation has emerged as a crucial visual generation task, aiming at synthesizing 3D content from textual descriptions. Studies of this task are currently shifting from per-scene T23D, which requires optimization of the model for every content generated, to General T23D (GT23D), which requires only one pre-trained model to generate different content without re-optimization, for more generalized and efficient 3D generation. Despite notable advancements, GT23D is severely bottlenecked by two interconnected challenges: the lack of high-quality, large-scale training data and the prevalence of evaluation metrics that overlook intrinsic 3D properties. Existing datasets often suffer from incomplete annotations, noisy organization, and inconsistent quality, while current evaluations rely heavily on 2D image-text similarity or scoring, failing to thoroughly assess 3D geometric integrity and semantic relevance. 
To address these fundamental gaps, we introduce GT23D-Bench, the first comprehensive benchmark specifically designed for GT23D training and evaluation.
We first construct a high-quality dataset of 400K 3D assets, featuring diverse visual annotations (70M+ visual samples) and multi-granularity hierarchical captions (1M+ descriptions) to foster robust semantic learning. Second, we propose a comprehensive evaluation suite with 10 metrics assessing both text-3D alignment and 3D visual quality at multiple levels. Crucially, we demonstrate through rigorous experiments that our proposed metrics exhibit significantly higher correlation with human judgment compared to existing methods. Our in-depth analysis of eight leading GT23D models using this benchmark provides the community with critical insights into current model capabilities and their shared failure modes. GT23D-Bench will be publicly available to facilitate rigorous and reproducible research.

\end{abstract}


\section{Introduction}
\label{sec:intro}
\IEEEPARstart{W}{ith} the rapid proliferation of digital 3D assets that play a critical role across architecture, gaming, and VR/AR, research on automatic 3D generation has attracted increasing attention in recent years~\cite{3DGen,Shap-E,MVDream}. Within the domain of automatic 3D generation, \textbf{text-to-3D (T23D)} generation has emerged as a particularly important task, as it enables the intuitive and accessible synthesis of 3D assets directly from natural language prompts.
Despite the significant advances in vision generation, T23D remains an extremely challenging task rooted in its intrinsic properties, such as text–3D alignment and 3D visual quality. Such a task can be viewed as a leap from the visual reconstruction toward true semantic understanding of the 3D world.

At the earlier stage, T23D methods typically optimized a scene-specific 3D representation (e.g., NeRF or 3DGS) for each input prompt through iterative refinement, which is referred to as per-scene T23D (PT23D). For example, ~\cite{DreamFusion, Magic3D, ProlificDreamer} adopt large-scale pre-trained text-to-image models as visual guidance and constrain the optimization of a 3D representation via Score Distillation Sampling (SDS). While PT23D achieves impressive results, its reliance on per-prompt optimization imposes significant computational and time costs, which in turn limit its generalization and scalability in practical applications. Recently, inspired by the successes of feed-forward models in vision generation tasks, generating diverse 3D content in a single feed-forward pass, namely general T23D (GT23D)~\cite{Point-E, Shap-E, dreamview, spad, VolumeDiffusion, MVDream, 3DTopia, cai2024semv}, has become a new research trend. In practice, such methods either lift pre-trained text-to-image models to generate multi-view representations~\cite{MVDream, dreamview, spad} or leverage the features of diffusion models as priors to directly construct 3D representations~\cite{Point-E, Shap-E, cai2024semv}, thereby avoiding per-prompt optimization. Consequently, GT23D substantially enhances inference efficiency and generalization, establishing itself as the prevailing paradigm for real-time and large-scale 3D generation.

\begin{figure*}
    \centering
    \includegraphics[width=\linewidth]{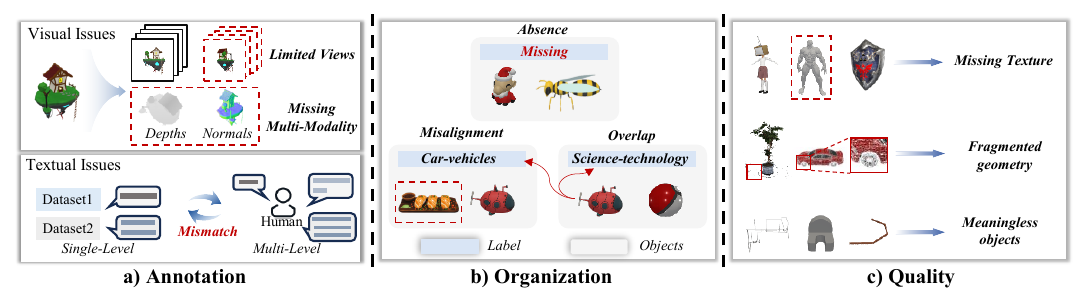}
    \vspace{-16pt}
    \caption{\textbf{Illustration of current GT23D dataset issues}: (a) annotation issues, including overly generic or overly detailed text and insufficient multi-modal or multi-view visual annotations; (b) organizational issues, such as absent, misaligned, or overlapping category labels; (c) asset quality issues, including missing textures, fragmented geometry, and non-semantic or abstract 3D objects. These issues collectively highlight the urgent need for a more structured, semantically aligned, and high-quality 3D dataset.}
    \label{fig:dataset_issue}
\end{figure*}

Nevertheless, progress in GT23D is still constrained by two interrelated challenges: the scarcity of high-quality, large-scale training data and the reliance on evaluation metrics that fail to account for intrinsic 3D characteristics.
In terms of data, existing 3D datasets~\cite{Objaverse, Cap3D} typically suffer from incomplete annotations, noisy organization, and inconsistent quality. Specifically, their visual annotations are incomplete across modalities and views, while textual descriptions are limited to single-granularity captions (Tab.~\ref{dataset_comp}). Moreover, noisy organization with mislabeled or overlapping categories (Fig.~\ref{fig:dataset_issue}-b) and degraded asset quality, such as fragmented or abstract shapes (Fig.~\ref{fig:dataset_issue}-c), collectively limit the utility of current 3D datasets.
Regarding evaluation, current assessments predominantly rely on 2D image–text similarity~\cite{CLIP} or GPT-based subjective scoring~\cite{t3bench,wu2024gpt}. Such metrics mainly measure overall semantic and visual quality at the appearance level, without adequately capturing fine-grained 3D semantic alignment or intrinsic geometric fidelity. Moreover, GPT-based judgments may inherit model biases and deviate from genuine human perception.
Taken together, these challenges in both data and metrics underscore that how to systematically facilitate large-scale pre-training and conduct robust, multi-dimensional evaluation for GT23D methods remains a pressing research imperative.

To this end, we introduce \textbf{GT23D-Bench}, the first comprehensive benchmark specifically designed for General Text-to-3D training and evaluation. We first construct 
\textbf{a high-quality 400K multimodal 3D dataset} through a three-stage pipeline of annotation, organization, and filtering. Specifically, each 3D asset is annotated with rich multimodal information, including multi-view RGB images, depths, normals (Fig.~\ref{fig:frontpage}.~A-i), and coarse-to-fine hierarchical textual descriptions (Fig.~\ref{fig:frontpage}.~A-ii) to enable diverse supervision signals. Then, following ImageNet’s~\cite{deng2009imagenet} structure, the dataset is hierarchically organized into seven primary categories and numerous semantically consistent subcategories based on WordNet~\cite{princeton_wordnet} (Fig.~\ref{fig:frontpage}.~A-iii), ensuring both diversity and structural clarity. Finally, we design a series of filtering algorithms specifically targeting different quality issues, ensuring both geometric completeness and visual fidelity.
Second, we propose \textbf{a comprehensive 3D-aware evaluation suite}, covering both text–3D alignment and 3D visual quality. The former measures the consistency between textual descriptions and 3D outputs at multiple granularities (Fig.~\ref{fig:frontpage}.~B-i), while the latter evaluates texture fidelity (e.g., using CC to assess sharpness of texture boundaries), geometric plausibility (e.g., Shape-C to detect fragments), and multi-view consistency (Fig.~\ref{fig:frontpage}.~B-ii). Extensive experiments also demonstrate that our 3D-aware metrics closely align with human perception.
\begin{figure}
    \vspace{-6pt}
    \centering
    \includegraphics[width=1.0\linewidth]{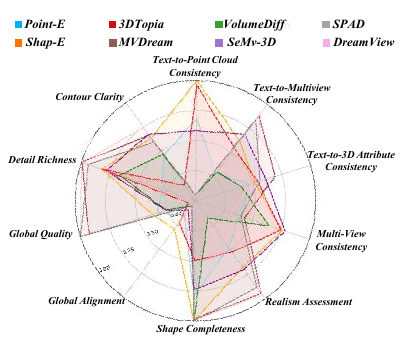}
    \caption{\textbf{Evaluation Results of Current GT23D Models based on GT23D-Bench Metrics.} We visualize the evaluation results of eight GT23D generation models in 10 dimensions. For comprehensive results, please refer to Tab.\ref{tab:rankings}}
    \label{fig:intro_radar}
\end{figure}
Thirdly, GT23D-Bench provides a comprehensive evaluation of current GT23D methods (Fig.~\ref{fig:intro_radar}), further \textbf{revealing several valuable insights}. (1) Existing methods effectively capture global semantics but struggle with fine-grained attribute alignment, revealing insufficient cross-modal supervision at detailed levels.
(2) The intrinsic biases of different 3D representations shape the balance between geometric accuracy and visual fidelity, indicating that representation learning remains a central bottleneck.
(3) The limited retention of image priors underscores the inherent difficulty of transferring 2D knowledge into 3D generative spaces.
(4) Current GT23D models remain predominantly perception-driven, prioritizing visual fidelity while hindering genuine spatial reasoning and structural comprehension.
(5) The tension between realism and imagination reflects an unresolved trade-off between fidelity and generalization.

Our contribution could be summarized as follows:
\begin{itemize}
    \item To the best of our knowledge, \textbf{GT23D-Bench} is the \textit{first} benchmark specifically tailored for GT23D, providing unified support for both large-scale pretraining and comprehensive evaluation.
    \item We construct a large-scale (400k+), high-quality text-to-3D dataset with rich multi-modal annotations (70M+) and hierarchical textual captions (1M+), and propose a comprehensive evaluation metrics (10) suite covering both text–3D alignment and 3D visual quality.
    \item  We perform a comprehensive evaluation and in-depth analysis of eight representative GT23D methods on GT23D-Bench, providing insights into their trade-offs between semantic alignment and geometric fidelity, and highlighting areas for future improvement. 
\end{itemize}

\begin{table*}
\centering
\caption{\textbf{Statistical comparison between existing GT23D datasets and our GT23D-Bench.} Compared to other datasets, our GT23D-Bench provides richer multimodal annotations for each object, including RGB images, depth maps, normal maps, and hierarchical captions, which demonstrates the comprehensive quality and strong advantages of our dataset. }
\label{dataset_comp}
\setlength{\tabcolsep}{0.5em}  
\renewcommand{\arraystretch}{1.2}  
\begin{tabular}{@{}clccccccc@{}} 
\toprule
\multicolumn{2}{@{}c}{\textbf{Dataset}} & 
\makecell{\textbf{Object} \\ \textbf{Quantity}} & 
\makecell{\textbf{Views} \\ \textbf{per Obj}} & 
\textbf{RGB} & 
\makecell{\textbf{Depth Map}} & 
\makecell{\textbf{Normal Map}} & 
\textbf{Caption} & \makecell{\textbf{Caption} \textbf{ Level}}\\ 
\midrule
\multirow{7}{*}{\makecell{Limited\\Scale\\(\textless{}100k)}} 
  & ShapeNetCore~\cite{ShapeNet} & 51k & -- & \checkmark & $\times$ & $\times$ & $\times$ & $\times$\\
  & ModelNet~\cite{ModelNet}     & 12k & -- & \checkmark & $\times$ & $\times$ & $\times$ & $\times$\\
  & 3DScan~\cite{3dscan}       & 2k  & $\sim$2k & \checkmark & $\times$ & $\times$ & $\times$ & $\times$\\
  & AMT Objects~\cite{AMTObjects}  & 2k  & -- & \checkmark & $\times$ & $\times$ & $\times$ & $\times$\\
  
  & Objectron~\cite{Objectron}    & 15k & $\sim$300 & \checkmark & $\times$ & $\times$ & $\times$& $\times$ \\
  & CO3D~\cite{CO3D}         & 19k & $\sim$200 & \checkmark & $\times$ & $\times$ & $\times$ & $\times$\\ 
  & GSO~\cite{GSO}          & 1k  & 5  & \checkmark & $\times$ & $\times$ & $\times$ & $\times$\\
  & Omniobject3D~\cite{omniobject3d}          & 6k  &   & \checkmark & $\times$ & $\times$ & $\times$ & $\times$\\
\midrule
\multirow{5}{*}{\makecell{Large\\Scale\\(\textgreater{}100k)}}  
  & MVImgNet~\cite{MVImgNet}     & 220k & $\sim$32 & \checkmark & \checkmark & $\times$ & $\times$ & $\times$\\
  & Objaverse~\cite{Objaverse}    & 818k & -- & $\times$ & $\times$ & $\times$ & $\times$ & $\times$\\
  & Cap3D~\cite{Cap3D}        & 661k & 20 & \checkmark & $\times$ & $\times$ & \checkmark & Coarse \\
  & 3DTopia~\cite{3DTopia}      & 120k & -- & \checkmark & $\times$ & $\times$ & \checkmark & Detailed \\
  \rowcolor{mycolor}& Ours         & 400k & 64 & \checkmark & \checkmark & \checkmark & \checkmark & Hierarchical \\
\bottomrule
\end{tabular}
\end{table*}

\section{Related Works}
\subsection{\textbf{Text-to-3D Generation}}
Text-to-3D generation~\cite{DreamFusion, ProlificDreamer, Magic3D, lucidreamer, MVDream, pami-t23dscene, pami-t23dshape} is a task that aims to synthesize 3D representations (3D voxels, point clouds, multiview images, and meshes) from textual descriptions. Depending on their generalization capability, existing approaches can be categorized into Per-scene Text-to-3D and General Text-to-3D methods.
\textbf{Per-scene Text-to-3D} generation has primarily been advanced by optimization-based methods like NeRF~\cite{NeRF} and 3DGS~\cite{3DGS}, but lacks zero-shot generalization. DreamFusion~\cite{DreamFusion} pioneered the use of pretrained text-to-image (T2I) diffusion priors through Score Distillation Sampling (SDS) to iteratively optimize 3D representations. Subsequent methods, such as ProlificDreamer~\cite{ProlificDreamer} and LucidDreamer~\cite{lucidreamer}, improve upon SDS via variational and interval formulations to enhance diversity and mitigate the Janus (multi-face) artifact. To enforce multi-view consistency, approaches like Magic123~\cite{Magic123}, RichDreamer~\cite{RichDreamer}, and ImageDream~\cite{ImageDream} fine-tune diffusion models toward 3D-aware generation. Meanwhile, methods such as Fantasia3D~\cite{Fantasia3D}, and Magic3D~\cite{Magic3D} explore explicit or hybrid representations, and amortized distillation frameworks~\cite{ATT3D,AToM} improve computational efficiency. Despite their ability to produce high-quality textures, these optimization-based pipelines remain computationally expensive—typically requiring several minutes to hours per instance. In contrast, General Text-to-3D approaches adopt a feed-forward paradigm, aiming for efficient, scalable, and generalizable 3D generation without per-scene optimization.

\textbf{General Text-to-3D} methods enable feed-forward 3D generation without per-scene optimization. According to their implementation strategies, they can be broadly classified into native-3D-based and multi-view-based approaches.
For native-3D-based methods, SDFusion~\cite{SDFusion} employs dense signed distance fields (SDFs) as 3D representations, which are computationally expensive and cannot render textures. Point-E~\cite{Point-E} and Shap-E~\cite{Shap-E} leverage large-scale 3D datasets to generate point clouds and meshes, respectively. 3DGen~\cite{3DGen} integrates a triplane variational autoencoder for learning latent representations of textured meshes with a conditional diffusion model that generates corresponding triplane features. VolumeDiffusion~\cite{VolumeDiffusion} enhances data efficiency by training a volumetric encoder to produce compact supervision for diffusion training. Given the scarcity of high-quality 3D data, recent works~\cite{cai2024semv} increasingly incorporate 3D priors to facilitate model learning.
For multi-view-based approaches, inspired by image-to-3D methods~\cite{Zero-1-to-3,One-2-3-45}, image diffusion models are adapted for 3D generation. MVDream~\cite{MVDream} and DreamView~\cite{dreamview} jointly train diffusion models on high-quality single-view and limited multi-view datasets to improve cross-view consistency and visual fidelity. Building upon MVDream, SPAD~\cite{spad} extends the capability to arbitrary-view generation.
Despite these advancements, existing methods still struggle to achieve strong text–3D semantic alignment while maintaining high 3D visual fidelity, highlighting the need for improved 3D-aware supervision and evaluation frameworks.

\subsection{\textbf{Text-to-3D Benchmarks}}
The benchmarks consist of two main components: datasets and evaluation metrics.
\textbf{In terms of data}, to systematically analyze existing available datasets for Text-to-3D, we categorize them into Limited-Scale datasets ($\le$100K objects) and Large-Scale datasets ($>$100K objects) based on their object quantity. Representative statistics are summarized in Tab.~\ref{dataset_comp}. For limited-scale datasets, early 3D datasets like ShapeNet~\cite{ShapeNet} and ModelNet~\cite{ModelNet} are two object-centric datasets offering extensive 3D Computer-Aided Design (CAD) models for early shape analysis. However, these models are often low-quality, untextured, and exhibit a significant domain gap from real-world objects. Later 3D datasets such as 3DScan~\cite{3dscan}, AMT Objects~\cite{AMTObjects}, and CO3D~\cite{CO3D} provide real-world scans but are still limited in quality (e.g., sparse point clouds) and exhibit constrained category distributions, restricting their generalizability for diverse 3D generation tasks.
While GSO~\cite{GSO} and OmniObject3D~\cite{omniobject3d} make efforts to provide high-quality datasets with category distributions more aligned with the real world, their limited scale—at most 6,000 objects—remains insufficient for comprehensive training in general 3D generation. In terms of large-scale datasets, recently, Objaverse~\cite{Objaverse} introduced a large-scale dataset comprising numerous richly textured 3D objects. However, its web-sourced nature results in coarse or missing annotations, inconsistent quality, and labeling errors. While Cap3D~\cite{Cap3D} and 3DTopia~\cite{3DTopia} enhance Objaverse with textual captions and Objaverse-XL~\cite{Objaverse-xl} expands it to 10 million objects, these challenges remain unresolved. 
Therefore, a more comprehensive, well-annotated, meticulously organized, and high-quality text-to-3D dataset is crucial for advancing text-to-3D generation.

Moving on to the \textbf{evaluation methods} of the benchmarks, evaluating text-guided generative models is challenging, yet crucial for assessing their generative capabilities.
For text-to-image (T2I) generation, early evaluations rely on distribution-based metrics such as FID~\cite{fid} and IS~\cite{salimans2016improved}, which assess visual realism against reference images. However, these metrics lose validity in open-domain settings where no ground truth exists. Consequently, recent benchmarks~\cite{pami-ganbench, huang2023t2i, pami-t2ibench,lu2024llmscore, li2024genai} incorporate large multimodal models~\cite{openai2023gpt4} and vision-language tools~\cite{li2022blip, zhou2022simple} to enable semantic- and alignment-aware assessments.
Similarly, early text-to-video (T2V) methods employ FVD~\cite{fvd}, a temporal extension of FID, to quantify visual quality against reference sequences. Modern benchmarks~\cite{huang2024vbench, ji2024t2vbench, liu2024evalcrafter, pami-vbench++, liu2024fetv} further adopt pretrained perceptual models to measure motion realism, temporal coherence, and prompt consistency, providing a more holistic evaluation protocol.
In text-to-3D (T23D) generation, evaluations have mostly extended 2D metrics (e.g., CLIP Score~\cite{CLIP}, Aesthetic Score~\cite{aesthetic}) to rendered images, enabling partial assessment of semantic and perceptual quality. Recent benchmarks like GPTEval3D~\cite{wu2024gpt} and T3Bench~\cite{t3bench} attempt to assess overall text-3D alignment and geometric quality, marking a step toward holistic evaluation of 3D generation. However, these benchmarks are primarily designed for PT23D methods that depend on fine-grained mesh reconstructions, rendering them less compatible with GT23D evaluation. In addition, their GPT-based scoring inherently carries model-induced biases, leading to subjective and potentially unreliable assessments. Consequently, a dedicated evaluation benchmark specifically tailored to the characteristics and objectives of GT23D is urgently needed.



\section{The Proposed GT23D-Bench}
In light of the aforementioned challenges in GT23D dataset and evaluation, we propose \textbf{GT23D-Bench}, the first comprehensive benchmark featuring a high-quality refined multimodal dataset (Sec.~\ref{sec:dataset}) and a suite of 3D-aware evaluation metrics (Sec.~\ref{sec:metrics}) for reliable and holistic pretraining and assessment of general text-to-3D generation.

\begin{figure*}
    \centering
    \includegraphics[width=\linewidth]{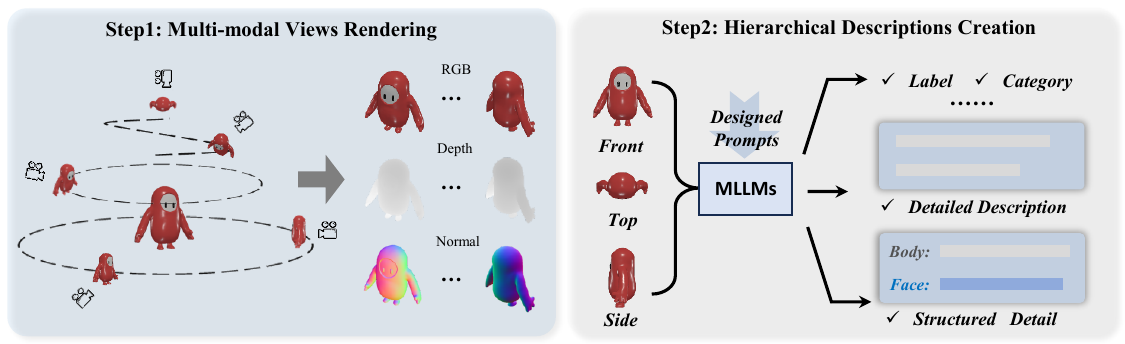}
    \caption{\textbf{Illustration of Dataset Annotation Pipeline.} 1) We render each object from 64 uniformly distributed viewpoints to obtain RGB images, depth maps, and normal maps by leveraging Pyvista. 2) We aggregate visual features from the front, top, and side views and use a large multimodal model to generate hierarchical textual annotations, ranging from coarse-grained category labels to fine-grained attribute-level descriptions. The more detailed design of the hierarchical description prompts is provided in Suppl.~I-C.}
    \label{fig:dataset_annotate}
\end{figure*}

\begin{figure}
    \centering
    \includegraphics[width=\linewidth]{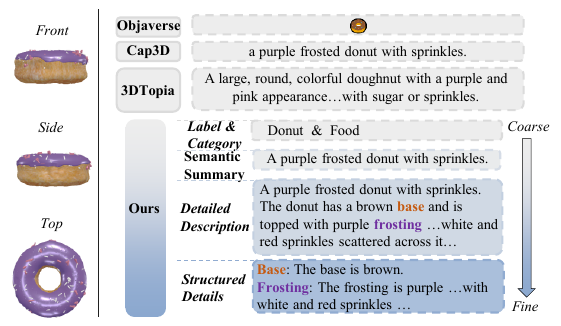}
    \caption{\textbf{Illustration of caption granularity differences across datasets.} Existing text–3D datasets typically provide captions of a single granularity—either overly simple or excessively detailed. In contrast, our hierarchical captions deliver more comprehensive and balanced object descriptions, integrating global semantics with fine-grained attribute details. }
    \label{fig:caption_case}
\end{figure}

\begin{figure}
    \centering
    \vspace{-12pt}
    \includegraphics[width=0.75\linewidth]{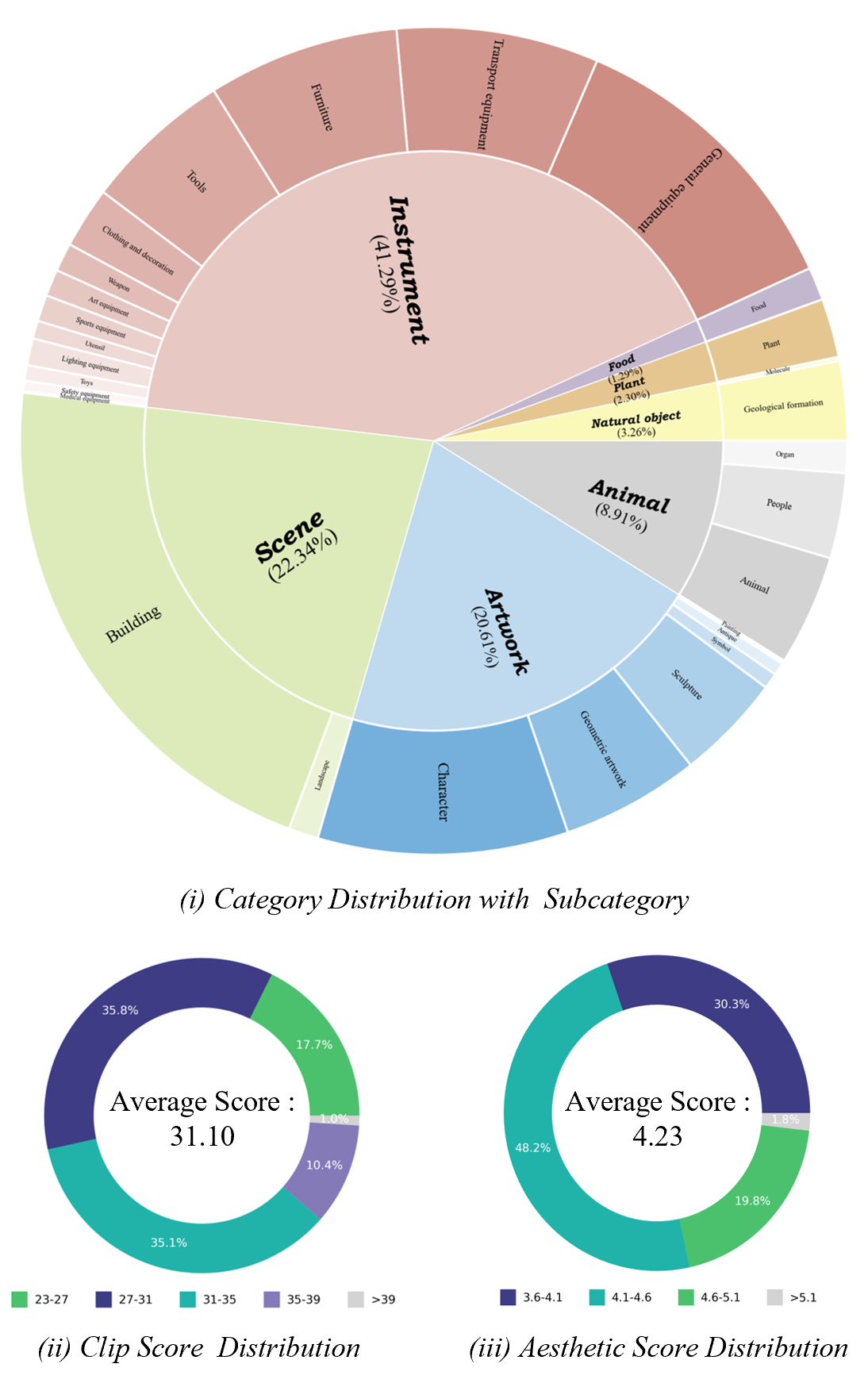}
    \vspace{-8pt}
    \caption{\textbf{Statistical overview of the GT23D-Bench dataset.} (i) Subcategory distribution visualization showing the overall composition of our dataset. (ii) CLIP score distribution representing the semantic diversity. (iii) Aesthetic score distribution reflecting the visual diversity.}
    \label{fig:data_dis}
\end{figure}


\subsection{\textbf{High-quality Refined Multimodal GT23D Dataset}}
\label{sec:dataset}
In response to the limitations observed in existing datasets, we introduce a systematic dataset construction pipeline consisting of three key stages: annotation, reorganization, and quality control. These stages are designed to ensure both structural consistency and data reliability, as detailed in the following subsections.

\subsubsection{\underline{Rich Annotation}}  
\noindent{\textbf{Multi-Visual Annotation}}:
We provide Objaverse~\cite{Objaverse} with multimodal annotations to facilitate paired data supervision in GT23D training and evaluation. 
As illustrated in Fig.~\ref{fig:dataset_annotate}, we uniformly sample \cite{pyvista} 64 multi-view RGB images by rotating around the azimuth angle at 360 degrees within three elevation orbits: 0 degrees, 30 degrees, and 60 to 90 degrees. We further include depth maps, normal maps, and camera parameters for each view, as previous works~\cite{instantmesh} have shown that multi-modal visual signals effectively improve 3D representation learning, facilitating more accurate and geometry-aware understanding.
\noindent{\textbf{Hierarchical Captioning}}:
Beyond enriched visual annotations, we further introduce hierarchical textual captions to enhance models’ robustness in semantic understanding.
For caption annotation, we adopt a multi-level strategy to capture varying levels of descriptive detail. Specifically, we employ the pre-trained multimodal model LLaVA~\cite{LLaVA} to caption a 3D object from its rendered three views(front, side, and top) with increasing levels of detail. As shown in Fig.~\ref{fig:caption_case}, each 3D object is annotated with a Label, a Semantic Summary, and a Detailed Description that characterizes part-level attributes and inter-part relationships from multiple viewpoints. In addition, we construct Structured Details by leveraging the pre-trained large language model LLaMA~\cite{touvron2023llama}, which transforms detailed descriptions into dictionary-style entries (e.g., {Base: The base is brown ...}) for more precise and machine-readable component-level understanding. More details on caption generation procedures are provided in \textbf{Suppl.~I-A}.

Human Alignment Comparison. To demonstrate that captions in our dataset better reflect human expressions, we conduct a user study comparing different caption methods on Objaverse, as shown in the first row (\textit{``Human Alignment"}) of Tab. \ref{tab:caption}. We invite 40 users to score caption quality from 1 to 5, collecting a total of 1600 scores. As shown in Tab. \ref{tab:caption}, our method significantly outperforms other caption methods on Objaverse in terms of human alignment.



\subsubsection{\underline{Object Reorganization}}




To address the dataset disorganization caused by label misalignment, label absence, and semantic overlap (with two labels), we implement a systematic approach during the Label-based Organization stage.


To ensure accurate label semantics, we first use the large language model LLaMA~\cite{touvron2023llama} to extract the main subject from each generated caption (e.g., “dinosaur” from “A green dinosaur with a gray pedestal”). We then classify these labels into seven primary categories—scene, instrument, plant, food, animal, artwork, and natural object—designed to align with ImageNet’s structure~\cite{deng2009imagenet}. For each category, we identify subcategories using WordNet~\cite{princeton_wordnet} and select appropriate sub-nodes.
Following this hierarchy, we assign each label to a primary category and its relevant subcategory, leveraging LLaMA~\cite{touvron2023llama} to generate new subcategories as needed, ensuring comprehensive dataset organization. More details on re-organization process are presented in \textbf{Suppl.~I-B}.
The final category distribution after the reorganization process is shown in Fig.~\ref{fig:data_dis} (i), while the subclass distribution after reorganization still follows a long-tail pattern consistent with real-world data, as shown in Fig.~\ref{fig:frontpage} (c).



\subsubsection{\underline{Quality Control}}
Since Objaverse is collected from the Internet, it suffers from issues such as fragmented objects, low scan quality, missing textures, and meaningless 3D shapes. To address these problems, we design a series of filtering algorithms:
\begin{figure*}
    \centering
    \includegraphics[width=0.9\linewidth]{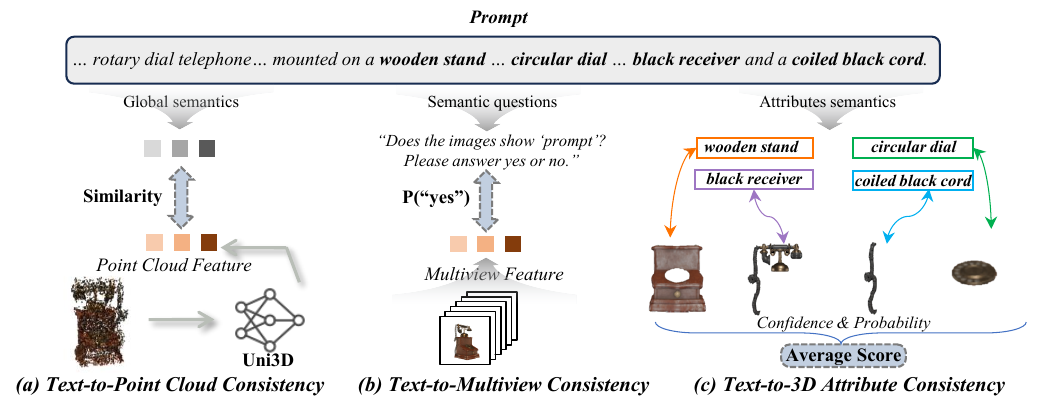}
    \vspace{-5pt}
    \caption{\textbf{Illustration of Text-3D Alignment Metrics.} We assess the alignment between textual descriptions and different levels of 3D representation, ranging from high-level semantics to detailed attributes: (a) point cloud–level evaluation measuring global semantic consistency, (b) multi-view–level evaluation assessing cross-view textual alignment, and (c) attribute–level evaluation capturing fine-grained text–3D correspondence.}
    \label{fig:metric-t-3d}
    \vspace{-8pt}
\end{figure*}
To remove fragmented objects, we perform connected component analysis on each rendered sample. An object is flagged as noisy and discarded if the cumulative area of isolated components smaller than 10 pixels exceeds 1\% of the object’s total area. We then eliminate objects lacking texture or exhibiting failed renders by detecting images with uniformly gray RGB values. To further refine data quality, we utilize the multimodal model LLaVA~\cite{LLaVA16} to identify and exclude unrecognizable or non-semantic objects by evaluating their multi-view images and captions for real-world relevance. Finally, we compute both the CLIP Score and Aesthetic Score for each object, removing those below the bottom 20\% and 30\% thresholds, respectively, to ensure high textual-visual alignment and visual quality. The statistics are illustrated in Fig.~\ref{fig:data_dis} (ii) and (iii). More details on the quality control process are provided in \textbf{Suppl.~I-C}.


\subsection{\textbf{Comprehensive 3D-Aware Evaluation Metrics}}
\label{sec:metrics}

Given the limitations in current GT23D evaluation, we argue for a rethinking of evaluation protocols: rather than solely extending scene-specific mesh benchmarks or relying on 2D proxies, there is a need for a \textbf{more general, geometry-aware, and human-aligned evaluation framework} that can accommodate diverse output modalities and provide a holistic assessment of GT23D generation quality.

Accordingly, we divide the evaluation of GT23D into two key components: \textbf{Text-3D Alignment Metrics} (Section~\ref{sec:Textual-3D Alignment}) and \textbf{3D-Visual Quality Metrics} (Section~\ref{sec:3D Visual Quality}), which together provide a more holistic understanding of the GT23D generation performance. More implementation details are provided in \textbf{Suppl.~II}.

During the evaluation, given a textual description \(\mathcal{T}\) and a GT23D generation model \( G \), we obtain the generated 3D representation \( G(\mathcal{T}) \). Then we render an image \(I\) every 30° and obtain a total of 12 multiview images \(I_{mv}\), and extract the corresponding point clouds \( P \). The evaluation is then carried out based on the following criteria.

\subsubsection{\underline{Text-3D Alignment Metrics}}
\label{sec:Textual-3D Alignment}
To thoroughly evaluate semantic consistency, we assess the alignment between textual descriptions and various levels of 3D representation, ranging from high-level semantics to detailed features. Specifically, we measure text-to-point-cloud consistency, text-to-multi-view consistency, and text-to-3D attribute consistency.


\noindent{\textbf{Text to Point Cloud Consistency (TP).}}
To evaluate semantic alignment from a global 3D perspective, we leverage the 3D representations produced by GT23D. However, unlike per-scene text-to-3D methods, GT23D is optimized for multi-view image synthesis, which often results in coarse and noisy 3D structures. Directly employing such representations for semantic evaluation may therefore introduce substantial noise.

To mitigate this issue and ensure a more stable assessment, we extract 3D visual features instead of relying on raw geometry. Specifically, we convert all 3D outputs into a unified point cloud representation $P$. We then employ the open-domain text–point-cloud model Uni3D~\cite{uni3d} to extract visual features from the point cloud $P$, and compute their similarity to textual features derived from the prompt $\mathcal{T}$, formulated as:
\begin{equation}
    Score_{TP} = Uni3D(\mathcal{T}, P).
\end{equation}

\noindent{\textbf{Text to Multi-Views Consistency (TV).}}
In order to further assess semantic consistency at a finer level, we turn to the multi-view perspective analysis. Although the Clip Score is designed to assess textual-image alignment, it primarily evaluates global alignment, often overlooking fine-grained details. To overcome this limitation, we employ VQAScore~\cite{VQAScore}, which is more sensitive to detailed and compositional text prompts. Given a text prompt \(\mathcal{T}\) and a set of multi-view rendered images \(I_{mv}\), VQAScore computes the Text to Multi-view Consistency as:

\begin{equation}
    Score_{TV} = VQAScore(\mathcal{T}, I_{mv}).
\end{equation}






\begin{figure*}
    \centering
    \includegraphics[width=0.95\linewidth]{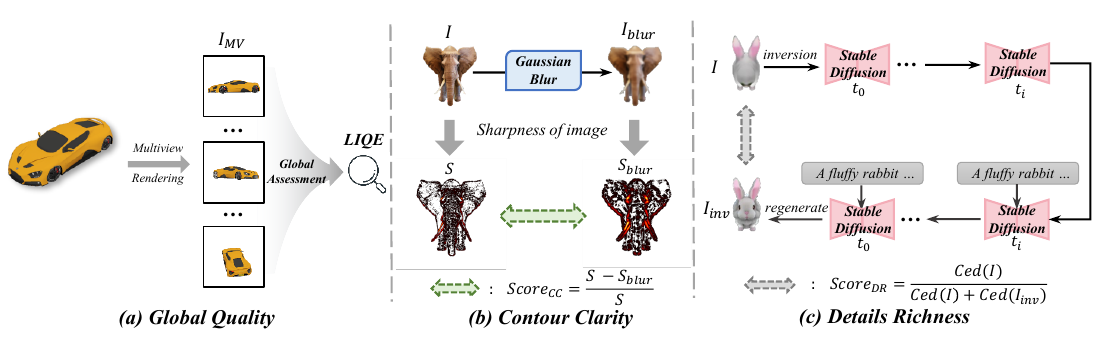}
    \vspace{-4pt}
    \caption{\textbf{Illustration of Texture Fidelity Metrics.} We employ a hierarchical evaluation strategy to provide a comprehensive assessment of both global perceptual fidelity and local textural quality: (a) overall visual realism of rendered textures, (b) texture sharpness capturing fine-grained detail clarity, and (c) texture richness evaluating color and pattern diversity.}
    \label{fig:metric-TF}
    \vspace{-8pt}
\end{figure*}
\noindent{\textbf{Text to 3D Attribute Consistency (TA).}} 
To assess the fine-grained semantic alignment of generated results, we conduct a component-level analysis from a localized attribute perspective. As illustrated in Fig.~\ref{fig:metric-t-3d} (c), we leverage attribute terms in the input prompts to identify and match corresponding object components across multiple views, using Grounded-SAM \cite{grounded-sam} for attribute grounding and localization.
For each attribute $i$, we compute both the detection confidence ($conf_i$) and the attribute classification probability ($prob_i$), capturing the presence and semantic correctness of the detected component. These signals are jointly aggregated to produce the final alignment score $Score_{T\text{}A}$ across all $N$ attributes:
\begin{equation}
Score_{T\text{}A} = \frac{1}{N} \sum_{i=1}^{N} conf_i \times prob_i.
\end{equation}  
This score quantitatively reflects both the detectability and semantic consistency of localized attributes within the generated 3D content.


\subsubsection{\underline{3D-Visual Quality Metrics}}
\label{sec:3D Visual Quality}
We decompose and evaluate a 3D shape quality from two vital components: Texture Fidelity and Geometry Correctness. The texture encapsulates the surface appearance of a 3D object. Differently, geometry defines the overall structure and shape of the object.
Additionally, for the GT23D methods that only generate multi-view images, it is also crucial to evaluate the correspondence among different views. 



In evaluating \textbf{Texture Fidelity}, we adopt a coarse-to-fine strategy: we first assess the overall visual realism of the rendered textures, and then perform a more detailed analysis focusing on texture sharpness and richness. This multi-level evaluation allows us to capture both global perceptual quality and local textural details.

\noindent{\textbf{Texture Fidelity -- Global Quality (GQ)}}
assesses the overall perceptual quality of rendered textures across multiple views of a 3D shape, taking into account both aesthetic characteristics and consistency with real-world geometry. To quantify this, we apply LIQE \cite{liqe}, a powerful image quality assessment model, to evaluate the set of multi-view renderings $I_{MV}$:
\begin{equation}
    Score_{GQ} = LIQE(I_{MV}).
\end{equation}


\noindent{\textbf{Texture Fidelity -- Contour Clarity (CC)} }assesses the clarity and definition of object contours and edges, determining the presence of blurring or noise in the generated 3D model.

As shown in Fig.~\ref{fig:metric-TF} (b), to evaluate the clarity of generated objects' views, we propose a re-blur evaluation strategy inspired by super-resolution. In this approach, we apply a Gaussian blur filter to the original view $I$, producing a blurred version $ I_{\text{blur}}$. The assessment score is based on the degree of change in sharpness between \( I \) and \( I_{\text{blur}} \). 
To quantify the sharpness, we measure the variation in pixel intensity gradients across the image for \( I \) and \( I_{\text{blur}} \), which are respectively denoted as \(S\) and \(S_{blur}\). 
By comparing the sharpness of \( I \) and \( I_{\text{blur}} \) and normalizing this difference, we obtain a final clarity score that reflects the fidelity of the generated image:
\begin{equation}
    Score_{CC} = \frac{S - S_{blur}}{S}.
\end{equation}

\begin{figure*}
    \centering
    \includegraphics[width=0.95\linewidth]{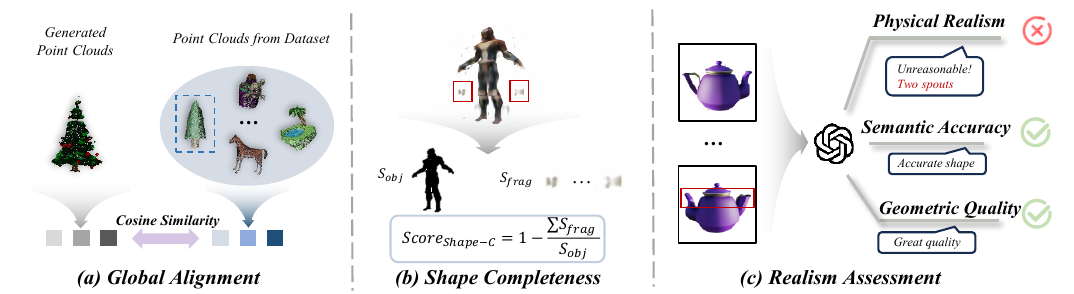}
    \vspace{-4pt}
    \caption{\textbf{Illustration of Geometry Correctness Metrics.} We evaluate geometric correctness across three complementary aspects reflecting the structural integrity and realism of generated 3D shapes: (a) global alignment, assessing the overall geometric similarity to real-world distributions; (b) shape completeness, measuring the integrity of the 3D shape; and (c) realism assessment, evaluating whether the overall structure is physically plausible and visually coherent. }
    \label{fig:metric-GC}
\end{figure*}

\noindent{\textbf{Texture Fidelity -- Details Richness (DR)}} evaluates how well the textures capture details and realism.
To determine whether an object is texture-rich, we leverage the powerful capabilities of the pre-trained image generation model, Stable Diffusion \cite{stable-diffusion}. As illustrated in Fig.~\ref{fig:metric-TF} (c), by applying an inversion process \cite{inversion}, we map a rendered view \(I\) back into the noise space, allowing Stable Diffusion to regenerate the object based on the inverted noise. 

\begin{figure}
    \centering
    \vspace{-8pt}
    \includegraphics[width=0.85\linewidth]{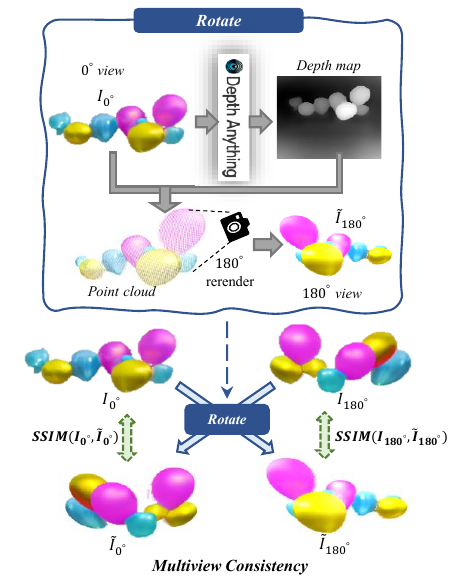}
    \vspace{-4pt}
    \caption{\textbf{Illustration of Multi-View Consistency (MV-Consis) Metric.}  
    We predict the depth map of the original view \( I_{0^\circ} \) and project it into 3D space to obtain a point cloud representation. 
    This point cloud is then reprojected into the target view at \(180^\circ\) to generate \( \tilde{I}_{180^\circ} \). 
    The visual similarity between \( I_{180^\circ} \) and \( \tilde{I}_{180^\circ} \) is measured to evaluate the multi-view consistency of the generated 3D object.
    }
    \label{fig:metric-mvconsis}
\end{figure}
Subsequently, we evaluate detail richness by comparing the level of detail between the regenerated view \( I_{\text{inv}} \) and the original rendered view \( I \) of the 3D object. Specifically, we employ a canny edge density metric \(Ced(\cdot)\) to quantify texture complexity. The detail richness score is then defined as:
\begin{equation}
Score_{DR} = \frac{ Ced(I) }{Ced(I) + Ced(I_{\text{inv}})}.
\end{equation}
To evaluate \textbf{Geometry Correctness}, we focus on three core aspects that ensure the structural integrity and logical coherence of 3D shapes.

\noindent{\textbf{Geometry Correctness -- Global Alignment (GA)}}
evaluates whether the generated 3D representations align with real-world distributions, we measure the feature-space distance between generated and real 3D representations. Specifically, we employ Uni3D~\cite{uni3d}, a pre-trained cross-modal 3D-text alignment model, to extract 3D features for both generated point clouds $P_{g}$ and ground truth point clouds $P_{gt}$ from real objects and then compute their cosine similarity.
\begin{equation}
    Score_{GA} = Cosine(Uni3D(P_{g}), Uni3D(P_{gt}))
\end{equation}

\noindent{\textbf{Geometry Correctness -- Shape Completeness (Shape-C)}
verifies the integrity of the generated 3D shape.
Concretely, if two regions lack directly connected colored pixels, they are identified as separate connected components. Components with an area smaller than a predefined pixel threshold (e.g., 50 pixels) are regarded as fragments, denoted as \( S_{frag} \). The ratio between the total area of all fragments and the object’s total area \( S_{obj} \) reflects the degree of incompleteness. Accordingly, the final completeness score is defined as:
\begin{equation}
    Score_{Shape-C} = 1-\frac{\sum S_{frag}}{S_{obj}},
\end{equation}
where a higher score indicates better geometric integrity.

\noindent{}\textbf{Geometry Correctness -- Realism Assessment (RA).}
To assess the geometric quality and realism of 3D models, we prompt (detailed prompts are provided in \textbf{Suppl.~II}) GPT~\cite{openai2023gpt4} with multi-view rendered images and request scoring based on three criteria: physical realism, semantic accuracy, and geometric quality. 

First, for physical realism, GPT evaluates whether model structures are realistic and obey physical rules $Score_1$. Second, in alignment with the textual description (semantic accuracy), GPT assesses whether the model's shape intuitively reflects its described object class $Score_2$. Finally, for geometric quality, GPT evaluates 3D coherence by identifying issues such as fragmentation, and disjointed parts $Score_3$. 
Ultimately, the three scores are averaged to assess the model's geometric realism and quality:
\begin{equation}
    Score_{RA} = Average(Score_1, Score_2, Score_3)
\end{equation}


\noindent{\textbf{Multi-view Visual Consistency (MV-Consis).}}
For GT23D methods that generate only multi-view images, such as MVDream, which produces four distinct 3D views, evaluating generation quality and semantic alignment alone is insufficient. Such methods often struggle to maintain coherence across views, resulting in inconsistent 3D representations that may still obtain deceptively high scores under conventional image-based metrics. This inconsistency highlights the need for evaluation protocols that explicitly account for cross-view geometric consistency.

To address this issue and ensure a more equitable evaluation, we propose a metric for assessing multi-view consistency - \textbf{MV-Consis}.
As illustrated in Fig.~\ref{fig:metric-mvconsis}, we leverage the pre-trained model DepthAnything \cite{depthanything} to estimate depth maps for multi-view images. These depth maps enable us to back-project the source view image into 3D space to form a point cloud, which is subsequently re-rendered into the target view using its corresponding camera parameters, yielding the reprojected image \( I_{rerender} \). Although minor geometric or color information may be lost during reprojection, the essential color distributions and contour structures remain preserved. To evaluate cross-view similarity, we compare the re-rendered image \( I_{rerender} \) with the target view \( I_{tar} \) using the {Structural Similarity Index (SSIM)}~\cite{ssim}, which focuses on structural and perceptual alignment rather than raw pixel correspondence. This choice mitigates sensitivity to reprojection-induced pixel discrepancies and provides a more robust measure of multi-view coherence.
\begin{equation}
    Score_{MV-Consis} = SSIM(I_{rerender},I_{tar})
\end{equation}


\section{Experiments}
\label{sec:experiments}
In this section, we first introduce our experimental settings (Sec.~\ref{sec:prompt_constrcut}), and then conduct a series of experiments covering dataset validation (Sec.~\ref{sec:experiments_dataset}), metric validation (Sec.~\ref{sec:metric_validation}), comprehensive GT23D evaluation (Sec.~\ref{sec:dimension_evaluation}), and analysis for GT23D (Sec.~\ref{sec:analysis_gt23d}). These experiments aim to comprehensively validate our proposed benchmark, ensuring the credibility, robustness, and insightfulness of GT23D-Bench.

\subsection{\textbf{Experiment Setting up}}
\noindent{}\textbf{GT23D Baseline Methods.}
We adopt eight representative GT23D methods for evaluation, including native-3D-based approaches Point-E~\cite{Point-E}, Shap-E~\cite{Shap-E}, 3DTopia~\cite{3DTopia}, VolumeDiffusion~\cite{VolumeDiffusion}, SeMv-3D~\cite{cai2024semv} and multi-view-based approaches MVDream~\cite{MVDream}, DreamView~\cite{dreamview}, SPAD~\cite{spad}. 

\noindent{}\textbf{Prompt Suite Construction.}
\label{sec:prompt_constrcut}
To ensure comprehensive and balanced evaluation, we design a two-part test prompt suite covering both open-domain generation and ground-truth–measurable accuracy.
For \textbf{open-domain evaluation}, we use GPT-4~\cite{openai2023gpt4} to generate 200 prompts across nine tiers formed by three levels of \emph{object complexity} and \emph{conceptual creativity}. These prompts span from simple, realistic objects to imaginative and abstract compositions, enabling systematic assessment of generative generalization beyond predefined datasets.
For \textbf{ground-truth evaluation}, we sample 180 prompts uniformly from six categories in our GT23D-Bench dataset, which provides paired text–3D data, ensuring diverse coverage of object types and scenes for direct evaluation of geometric and semantic fidelity.
Together, these two sets provide a robust foundation for assessing both grounded accuracy and open-domain generative capability. Implementation details are provided in \textbf{Suppl.~III}.

\begin{table*}[]
\renewcommand\arraystretch{1.5}
\caption{\textbf{Comparison of Human Alignment with Current GT23D Evaluation Metrics in 10 dimensions.} Correlation coefficient Pearson's $\rho$, Spearman's $\rho$, and Kendall's $\tau$ are used to measure the correlation between human judgments and metric scores, demonstrating that our proposed metrics exhibit stronger alignment with human evaluations compared to existing metrics.}
\resizebox{1\linewidth}{!}{
\centering
\begin{tabular}{l|c|cccccccccc}
\toprule
    \multirow{2}*{\textbf{\textit{Correlation}}} & \multirow{2}*{\makecell{\textbf{\textit{Metric}} \textbf{\textit{Baselines}}}} &\multirow{2}*{\makecell{\textbf{Text to PC } \\ \textbf{Consistency}}} & \multirow{2}*{\makecell{\textbf{Text to MV} \\ \textbf{Consistency}}} & \multirow{2}*{\makecell{\textbf{Text to Attri} \\ \textbf{Consistency}}} & \multicolumn{3}{c}{\textbf{Texture Fidelity}} & \multicolumn{3}{c}{\textbf{Geometry Correctness}} & \multirow{2}*{\makecell{\textbf{Multi-View} \\ \textbf{Consistency}}} \\
    && & & & GQ  & CC  & DR	& GA  & SC	& RA & \\
    \cline{1-12}
\multirow{5}{*}{\textbf{Person}($\rho$) $\uparrow$}       
    &Clip Score~\cite{CLIP} & 0.2315 & 0.3032& 0.1792& 0.1899 &0.2088 &0.2509 & -0.0795 &0.0245 &0.0837 & 0.0868\\ 
    &Aesthetic Score~\cite{aesthetic} & 0.1979 & 0.5077 & 0.3910 & 0.0703 & 0.1486 & -0.0721 & 0.2421 & 0.1560 & 0.4177 & -0.0158\\ 
    &T3Bench~\cite{t3bench} & 0.1100 & 0.3280 & 0.3025 & 0.0650 & 0.1815 & -0.1697 & 0.1950 & 0.2294 & -0.0655 & 0.2338\\ 
    &GPTEval3D~\cite{wu2024gpt} & -0.0033& 0.3339 & 0.3165 & 0.0160 & 0.0758 & -0.2464 & 0.3307 & 0.1413 & 0.1474 & -0.0986\\ 
    &Ours & \textbf{0.3637} &\textbf{0.5190} & \textbf{0.5061} & \textbf{0.4413} &\textbf{ 0.4815} & \textbf{0.3659} & \textbf{0.5785} & \textbf{0.4046} & \textbf{0.7379} & \textbf{0.6061} \\ 
    \cline{1-12}
\multirow{5}{*}{\textbf{Spearsman}($\rho$) $\uparrow$}  
    &Clip Score~\cite{CLIP}& 0.2157 & 0.2693 & 0.0724 & 0.2041 & 0.1871 & 0.2815 & -0.0177 & 0.0635 & 0.0612 & 0.2585\\
    &Aesthetic Score~\cite{aesthetic}& 0.2910 & 0.4836 & 0.3842 & 0.1683 & 0.1171 & -0.0895 & 0.1807 & 0.0726 & 0.4005 & 0.0749 \\
    &T3Bench~\cite{t3bench} & 0.2138 & 0.3380 & 0.3416 & 0.0974 & 0.1276 & -0.0405 & 0.2016 & 0.2189 & -0.0215 & 0.2188\\ 
    &GPTEval3D~\cite{wu2024gpt} & 0.0210 & 0.3836 & 0.4084 & 0.0468 & 0.1845 & -0.2841 & 0.3819 & 0.1229 & 0.3396 & -0.0476\\ 
    &Ours & \textbf{0.2721} & \textbf{0.5597} &\textbf{0.4762} & \textbf{0.4804} & \textbf{0.5515} & \textbf{0.3090} & \textbf{0.5637} & \textbf{0.4222} & \textbf{0.6683} & \textbf{0.4518} \\
    \cline{1-12}
\multirow{5}{*}{\textbf{Kendall}($\tau$) $\uparrow$}    
    &Clip Score~\cite{CLIP} & 0.1402 & 0.2003 & 0.0488 & 0.1400 & 0.1387 & \textbf{0.2016} & 0.0197 & 0.0686 & 0.0160 & 0.1889 \\ 
    &Aesthetic Score~\cite{aesthetic} & 0.2081 & 0.3705 & 0.2830 & 0.1193 & 0.0892 & -0.0344 & 0.1084 & 0.0081 & 0.2937 & 0.0533 \\
    &T3Bench~\cite{t3bench} & \textbf{0.1921} & 0.2706 & 0.2985 & 0.0882 & 0.0892 & -0.0147 & 0.1479 & 0.1486 & -0.0267& 0.1017\\ 
    &GPTEval3D~\cite{wu2024gpt} & 0.0501 & 0.2616 & 0.2733 & 0.0001 & 0.0991 & -0.2016 & 0.2886 & 0.0709 & 0.2575 & -0.0438\\ 
    &Ours& \underline{0.1802} & \textbf{0.3905} & \textbf{0.3318} & \textbf{0.3475} & \textbf{0.4434} & \underline{0.1917} & \textbf{0.4338} & \textbf{0.3506} & \textbf{0.5538} & \textbf{0.3632}\\
\bottomrule
\end{tabular}}
\label{tab:metrics}
\end{table*}

\begin{figure}
    \centering
    \includegraphics[width=0.9\linewidth]{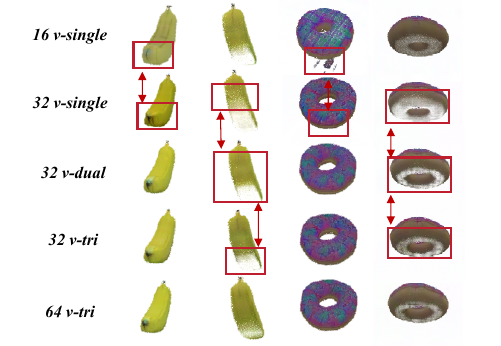}
    \vspace{-6pt}
    \caption{\textbf{Qualitative Analysis of Different View Quantity and Orbit Configuration in GT23D.} \textit{v} denotes the number of training views, and \textit{single/dual/tri} indicate the number of camera orbits used for rendering. We observe that increasing both the number of viewpoints and the angular coverage provided by multiple orbits leads to more stable and geometrically consistent 3D generation.}
    \label{fig:data-view}
\end{figure}

\begin{table}
\centering
\caption{\textbf{Quantitative Comparison of Caption Datasets with Varying Granularity.} The columns correspond to different levels of test prompt granularity (brief, normal, and detailed), while the rows represent different fine-tuning datasets. Models fine-tuned on our hierarchical caption dataset achieve the highest CLIP scores across all prompt granularities, demonstrating the effectiveness and robustness of hierarchical captioning.}
\label{tab:caption_ablation}
\begin{tabular}{lcccc}
\toprule
\textbf{Training Setting} & \textbf{Brief} & \textbf{Normal} & \textbf{Detailed} & \textbf{Average} \\
\midrule
Detailed-only             & 25.49          & 26.79           & \underline{32.26}         & 27.68            \\
Hierarchical (Ours)      & \textbf{30.79} & \textbf{31.88}  & \textbf{33.16}& \textbf{31.76}   \\
\bottomrule
\end{tabular}
\end{table}

\subsection{\textbf{Validity Evaluation of GT23D-Bench Dataset}}
\label{sec:experiments_dataset}
To validate the effectiveness of our dataset, we conduct comprehensive experiments to analyze its advantages, demonstrating how it contributes to improving GT23D generation. All validation finetuning studies adopt the same GT23D base method, SeMv-3D.

\noindent{\textbf{Effect of View Diversity.}}  
Fig.~\ref{fig:data-view} presents the results of several controlled finetuning experiments to investigate the impact of view diversity. Specifically, we examine five configurations: i) 16v-single: 16 views from a single orbit, ii) 32v-single: 32 views from a single orbit, iii) 32v-dual: 32 views from two orbits, iv) 32v-tri: 32 views from three orbits, and v) 64v-tri: 64 views from three orbits. Intuitively, within a fixed orbital range, increasing the number of rendered views yields more complete and geometrically consistent reconstructions, as evident from the comparison between 16v-single and 32v-single.
Moreover, when holding the number of views constant, expanding the orbital coverage further improves both geometric accuracy and textural fidelity. For instance, the 32v-tri results exhibit a more solid structure and fewer fragmented artifacts than 32v-dual or 32v-single.
In conclusion, the full multi-orbit, multi-view configuration (64v-tri) produces the most faithful and detailed reconstructions, validating the effectiveness of our dataset design and view annotation strategy.

\noindent{\textbf{Effect of Hierarchical Captions.}}
In Tab.~\ref{tab:caption_ablation}, we analyze the effect of caption granularity by finetuning the base model with either detailed-only captions (Detailed-only) or mixed multi-granularity captions (Hierarchical). To this end, we compute CLIP scores between rendered views of the generated 3D content and the text prompts of different lengths (brief, normal, detailed), evaluating how well the model preserves semantic alignment across varying levels of prompt granularity. Obviously, the Hierarchical model substantially outperforms the Detailed-only model, achieving an average gain of 14.7\% and improvements of 20.8\%, 19.0\%, and 2.8\% on brief, normal, and detailed prompts, respectively. Remarkably, the +2.8\% improvement on detailed prompts indicates that introducing coarse-to-fine caption diversity does not dilute fine-grained semantics but instead reinforces them. Taken together, these results demonstrate that hierarchical captioning significantly enhances both semantic fidelity and robustness, enabling the model to handle diverse linguistic expressions more reliably.

\noindent{\textbf{Effect of Re-organization and Quality Control.}}
To evaluate the effectiveness of our re-organization and quality control pipeline, we compare dataset statistics before and after curation. Specifically, we assess improvements using CLIP Score and Aesthetic Score, which respectively measure text–visual alignment and perceptual quality.
After curation, the average CLIP Score increases from 30.2 to 31.1, and the Aesthetic Score rises from 3.91 to 4.23, indicating clearer semantic correspondence and enhanced visual appeal. More qualitative results are provided in \textbf{Suppl. I}.
Overall, these results demonstrate that re-organization and quality control meaningfully improve both the semantic reliability and perceptual quality of the dataset, thereby providing a stronger foundation for subsequent GT23D training and evaluation.
\begin{table}[t]
    \centering
        \caption{\textbf{User Study on data Quality across Current GT23D Datasets.} We conduct a comparative study on Objaverse, Cap3D, 3DTopia, and Ours. Specifically, \textit{\textbf{Human Alignment}} measures human preference for the quality and semantic coherence of dataset captions, while \textit{\textbf{Training Boost}} reflects human preference for the visual and semantic quality of 3D generations produced by GT23D models trained with captions from different datasets.}
    \resizebox{\linewidth}{!}{\begin{tabular}{lcccc} \toprule
 &\textbf{Objaverse} & \textbf{Cap3D} & \textbf{3DTopia} & \textbf{Ours} \\ \midrule
\textbf{Human Alignment ↑} & 2.34 & 3.11 & 3.50 & \textbf{3.89}\\
\textbf{Training Boost ↑} & 2.81 & 3.68 & 3.85 & \textbf{4.28}\\
 \bottomrule
    \end{tabular}}
    \label{tab:caption}
\end{table}

\begin{figure*}
    \centering
    \includegraphics[width=0.9\linewidth]{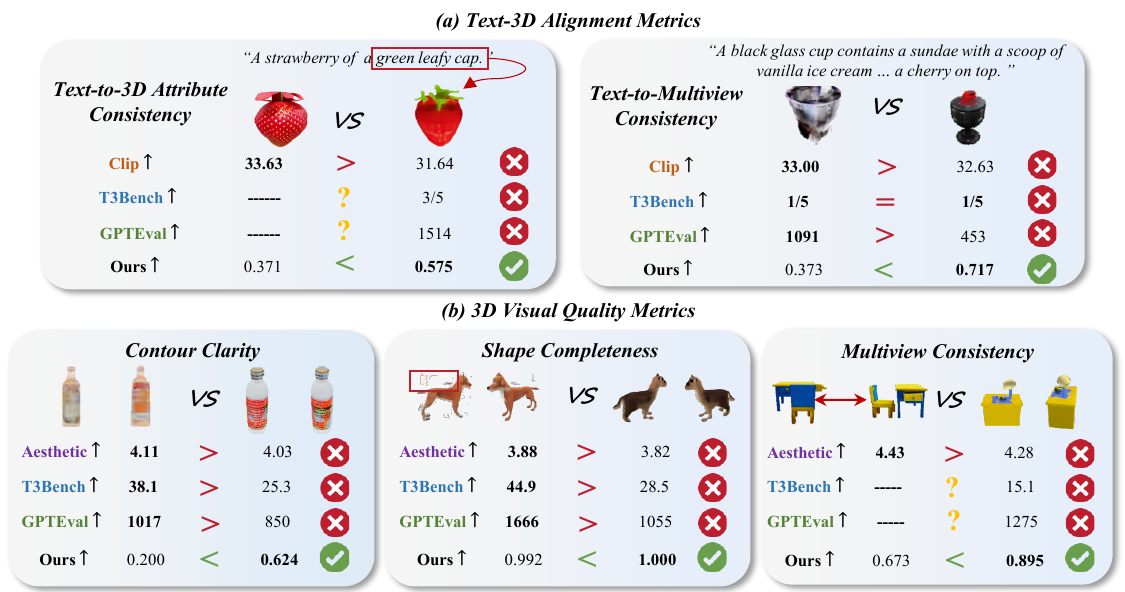}
    \vspace{-6pt}
    \caption{\textbf{Qualitative Comparison with Current GT23D Metrics.} We compare several evaluation metrics, including CLIP Score (Clip), Aesthetic Score (Aesthetic), T3Bench, GPTEval3D (GPTEval), and our proposed metric, across identical evaluation dimensions. For benchmarks incapable of assessing pure multi-view generation methods, the corresponding entries are denoted as “----”. The results show that our metric achieves higher consistency with the observed semantic alignment and visual quality in each dimension, underscoring its reliability and effectiveness.}
    \label{fig:baseline_com}
\end{figure*}

\noindent{}\textbf{Effect of Overall Training Boost.}
To verify whether our dataset more effectively supports GT23D model training, we compare the performance of the same base model after fine-tuning on different datasets. In practice, we fine-tune the base model on four representative datasets (Objaverse~\cite{Objaverse}, Cap3D~\cite{Cap3D}, 3DTopia~\cite{3DTopia}, and our dataset) and conduct a user study to evaluate the generated outputs under identical prompts, as reported in \textit{Training Boost} row of Table~\ref{tab:caption}. Notably, the model trained on our dataset attains the highest mean rating of 4.28, demonstrating that our data provides stronger supervision and yields 3D results with improved semantic consistency and perceptual fidelity.

\subsection{\textbf{Validity Evaluation of GT23D-Bench Metrics}}
\label{sec:metric_validation}
To demonstrate the validity and human alignment of our GT23D-Bench metrics, we analyze them from both quantitative and qualitative perspectives, including: (i) quantitative correlation studies that assesses their consistency with human judgments, (ii) qualitative comparisons demonstrating their advantages over existing benchmarks, and (iii) qualitative sensitivity analyses validating their responsiveness to perceptible quality variations.

\noindent{}\textbf{Quantitative Correlation with Human Judgments}.
To evaluate whether our proposed metrics align with human perception, we analyze the correlation between metric scores and human ratings of GT23D outputs.
Specifically, for human ratings, we conduct a user study on 8 representative GT23D baselines to assess their text-to-3D alignment and visual fidelity (details in \textbf{Suppl. IV}). Correspondingly, for metric scores, we evaluate the exact same set of 3D outputs using existing evaluation methods (including CLIP Score~\cite{CLIP}, Aesthetic Score~\cite{aesthetic}, T3Bench~\cite{t3bench}, and GPTEval3D~\cite{wu2024gpt}) and our proposed metrics. 
Finally, we measure Pearson’s $\rho$, Spearman’s $\rho$, and Kendall’s $\tau$ correlations between the collected human ratings and the calculated metric scores. For a fair comparison, correlations are computed only for dimension-matched metrics (e.g., T3Bench’s text-alignment sub-metric with Text-to-Multiview Consistency).
As summarized in Tab.~\ref{tab:metrics}, our proposed metrics consistently achieve the highest correlations across most dimensions, demonstrating superior alignment with human judgments in both semantic consistency and visual fidelity.

\begin{figure}
    \centering
    \includegraphics[width=0.9\linewidth]{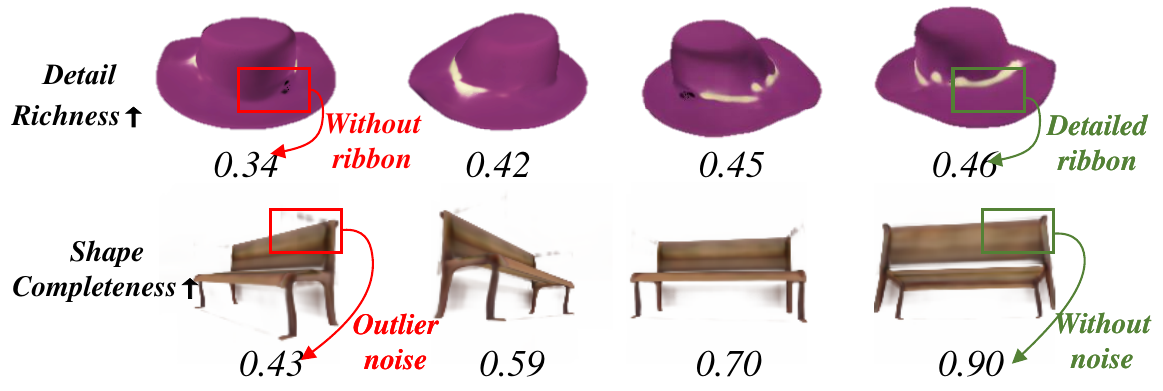}
    \vspace{-8pt}
    \caption{\textbf{Visual Illustration of Metric Sensitivity.} Visual examples demonstrating our metrics’ ability to reflect variations in “Detail Richness” and “Shape Completeness.”}
    \label{fig:score_vary}
\end{figure}

\begin{table*}[!htbp]
\renewcommand\arraystretch{1.5}
\centering
\caption{\textbf{Quantitative comparison of different GT23D baselines using GT23D-Bench metrics in 10 different evaluation dimensions.} The higher scores indicate better performance. Since some methods do not output explicit 3D representations, their scores are absent in TP and GA which need point clouds.}
\resizebox{0.95\linewidth}{!}{
\begin{tabular}{lcccccccccc}
\hline
           & TP     & TV     & TA     & GQ     & CC     & DR     & GA     & Shape-C      & RA & MV-Consis \\ \hline
Point-E~\cite{Point-E}    & 0.1964 & 0.3041 & 0.3518 & 1.8919 &    0.4314    & 0.4706       & 0.36   & 99.1669 &  0.5447  & 0.8423    \\
Shap-E~\cite{Shap-E}     & \textbf{0.3017} & 0.5543 & 0.3617 & 2.0165 & 0.5495 & 0.8276 & \textbf{0.4639} & \textbf{100}     &   0.6733 & 0.8974    \\
VolumeDiff~\cite{VolumeDiffusion} & -      & 0.3188 & 0.3364 &    1.6988    &  0.4399      &   0.7243     &    -    &     99.99    &  0.3567  &   0.8802        \\
3DTopia~\cite{3DTopia}    & 0.2559 &     0.4989   &  0.3673      &  1.1683      &  0.2322      &   0.8261     &   0.4122     &   97.9655      & 0.5753   &   0.8956        \\
SeMv-3D~\cite{cai2024semv}    & 0.1746 &    0.6068    &    0.3828    &  1.7031      &  0.5765      &   0.8     &  0.3138      &     98.9453    & 0.686   &    \textbf{0.9008}       \\
MVDream~\cite{MVDream}    & -      &   0.7510     &  \textbf{0.4007 }     &  3.4995      & 0.5707       &   0.9468     &   -     &     99.9578    &  0.84  & 0.8484          \\
DreamView~\cite{dreamview}  & -      &    \textbf{0.7717}    &   0.3574     &  \textbf{3.8768}      &   \textbf{0.6203}     &   \textbf{0.9727}     &   -     &   99.9805      &  \textbf{0.8533}  &  0.8302         \\
SPAD~\cite{spad}       & -      &   0.7138     &  0.3725      &  3.8419      &    0.5224    &    0.9082    &    -    &    99.9892     &  0.8033  &     0.8430      \\ \hline
Min        & 0.0498 &    0.0977    &  0.2507      &   1.001     &   0.1271     &  0.2724      &    0.2407    &   95.9823      & 0.2433   &     0.7839      \\
Max        & 0.3031  &     0.8358   &   0.4661     &   3.6995     &    0.7902    &  0.9490      &  1.000     &   100.00      &   0.8767 & 0.9326          \\ \hline
\end{tabular}

}

\label{tab:rankings}
\end{table*}

\noindent{}\textbf{Qualitative Comparison across Evaluation Dimensions}.
To further validate the superiority of our metrics from a visual perspective, we present qualitative comparisons with CLIP Score (CLIP), Aesthetic Score (Aesthetic), T3Bench, and GPTEval3D (GPTEval) in Fig.~\ref{fig:baseline_com}. As shown, our scores exhibit greater alignment with perceptual quality across different evaluation dimensions. For \textit{text–3D alignment}, CLIP assigns a higher score to the left strawberry due to its vivid appearance, yet fails to recognize that only the right strawberry corresponds to the textual attribute “green leafy cap” (Fig.~\ref{fig:baseline_com}a-i). Meanwhile, existing benchmarks such as T3Bench lack sufficient semantic granularity. As shown in Fig.~\ref{fig:baseline_com}a-ii, it assigns identical low scores (1/5) to both objects and fails to identify the semantically correct one on the right.
For \textit{3D visual quality}, conventional metrics are similarly biased toward surface-level realism. In the “Multiview Consistency” example (Fig.~\ref{fig:baseline_com}b-iii), the Aesthetic metric favors the left model for its visual appeal despite inconsistent views, whereas our metric accurately reflects this inconsistency. Likewise, in the “Shape Completeness” example (Fig.~\ref{fig:baseline_com}b-ii), GPTEval3D gives a high score to the left dog due to its photorealistic rendering while overlooking its structural incompleteness.
In contrast, our method directly evaluates the structural integrity of the 3D shape, yielding scores that more faithfully correspond to the actual geometric quality.
These results confirm that our GT23D-Bench metrics more accurately capture both semantic fidelity and structural integrity, providing a perceptually aligned and geometry-aware evaluation framework for GT23D.

\noindent{\textbf{Qualitative Analysis of Metric Sensitivity.}}
Moreover, to validate that our metrics are sensitive to variations in 3D generation quality, we provide qualitative results in Fig.~\ref{fig:score_vary} focusing on \emph{Detail Richness} and \emph{Shape Completeness}. As shown in the bottom row, when outlier noise is progressively reduced from left to right, the corresponding Shape-C scores increase in a consistent and monotonic manner. This trend confirms that the metric responds appropriately to improvements in geometric integrity and can reliably capture quality differences across generated 3D outputs.


\begin{figure*}
    \centering
    \includegraphics[width=0.9\linewidth]{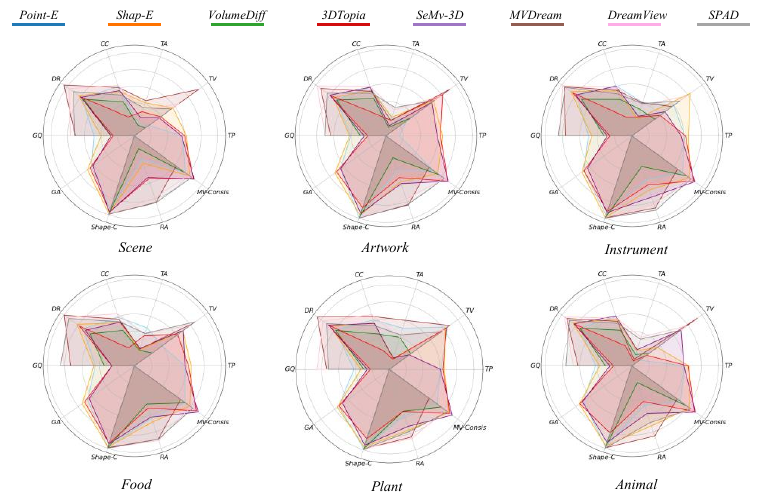}
    \vspace{-10pt}
    \caption{\textbf{Visualization of Quantitative Comparisons of Different GT23D baselines.} Radar plots showing the performance of existing GT23D baselines evaluated using GT23D-Bench metrics across six major categories. More qualitative comparison results are provided in Suppl. VI.}
    \label{fig:baseline_radar}
\end{figure*}
\subsection{\textbf{Comprehensive Evaluation of GT23D}}
\label{sec:dimension_evaluation}
\subsubsection{\underline{Per-Dimension Evaluation}}
For each evaluation dimension described in Sec.~\ref{sec:metrics}, we evaluate GT23D-Bench across different GT23D methods and report the results in Tab.~\ref{tab:rankings}. In addition, inspired by~\cite{huang2024vbench}, we introduce two reference baselines to approximate the empirical upper and lower bounds. 

\noindent{\textbf{Empirical Max.}} To approximate the upper bound for each metric, we randomly sample 2,000 cases from the Objaverse dataset and compute scores across most dimensions, taking the average of the top 5\% as the empirical maximum. For semantics-related metrics, we retrieve the corresponding ground-truth objects based on captions from the prompt suite and compute their scores as the theoretical upper bound. 

\noindent{\textbf{Empirical Min.}} For most metrics, we approximate the lower bound by evaluating randomly generated objects composed of Gaussian noise. For dimensions such as \textit{``multi-view consistency}, we instead construct view concatenations from ground-truth objects specifically designed to yield low scores. Further implementation details are provided in the Suppl.~V. 

\noindent{\textbf{Observations.}}
As shown in the table, multi-view–based methods generally achieve superior performance in texture fidelity and overall semantic alignment, in some cases even surpassing the empirical upper bound, such as DreamView on the DR metric ($0.9727 > 0.9490$). However, these methods tend to exhibit inferior multi-view consistency compared with native-3D–based approaches (as shown in the MV-Consis row). In contrast, native-3D–based methods achieve stronger geometric accuracy and cross-view coherence, albeit with lower texture quality. Overall, current GT23D methods demonstrate complementary advantages and limitations, suggesting that effectively balancing semantic alignment, texture fidelity, geometric correctness, and multi-view consistency remains an open and challenging research problem.

\subsubsection{\underline{Per-Category Evaluation}}
\label{sec:category_evaluation}
We evaluate GT23D methods across six primary categories defined in our dataset (excluding the "others" class). For each category, corresponding subsets from our \textit{prompt suite} in Sec.~\ref{sec:prompt_constrcut} are employed to assess the performance of different methods across multiple evaluation dimensions. The results, visualized in Fig.~\ref{fig:baseline_radar}, reveal distinct performance characteristics of each method under varying semantic and geometric conditions.

\noindent{\textbf{Observations.}}
We observe that for relatively common categories such as \textit{Instrument} and \textit{Food}, most methods perform reasonably well. However, when it comes to more complex categories like \textit{Artwork} or large-scale \textit{Scene}, native-3D–based methods exhibit a noticeable decline in performance. Meanwhile, although multi-view–based methods maintain relatively high semantic alignment and visual quality, they suffer a substantial drop in multi-view consistency. These results suggest that existing GT23D methods still struggle to handle complex and abstract 3D generation tasks that require coherent geometry and semantics across diverse spatial contexts.

\subsection{\textbf{Insights and Discussions}}  
\label{sec:analysis_gt23d}
In this section, we discuss the observations we draw from our comprehensive evaluation experiments. These analyses aim to uncover performance trends, identify current limitations, and provide insights to guide future research in GT23D.

\textbf{(1) Existing methods effectively capture global semantics but struggle with fine-grained attribute alignment, revealing insufficient cross-modal supervision at detailed levels.}
Current GT23D approaches can generally align generated 3D content with overall textual semantics but often fail to preserve fine-grained attributes, particularly in compositional or multi-object scenarios. Even advanced models such as MVDream, which leverage powerful pre-trained text-to-image priors, exhibit noticeable inconsistencies in component structures and appearances. This limitation likely stems from the scarcity of high-quality, fine-grained text–3D annotations. Our experiments (Tab.~\ref{tab:caption_ablation}) empirically validate this observation: models trained with hierarchical captions achieve substantially higher semantic accuracy, underscoring the importance of dataset granularity for robust GT23D performance.

\textbf{(2) The intrinsic biases of different 3D representations shape the balance between geometric accuracy and visual fidelity.}
Distinct 3D representations introduce specific biases in how textual semantics are translated into spatial and visual domains. Multi-view image–based models (e.g., MVDream, DreamView) inherit rich 2D priors, producing high-fidelity textures yet inconsistent geometry (high DR and GQ but low Shape-C and MV-Consis in Tab.~\ref{tab:rankings}). In contrast, explicit 3D formats such as point or voxel representations (e.g., Point-E) provide stronger geometric grounding (high TP) but often sacrifice visual realism. Implicit neural fields (e.g., Shap-E) demonstrate a more balanced capability across DR, TP, and MV-Consis by jointly encoding geometry and appearance in a continuous latent space. These findings suggest that implicit representations may offer a promising path toward reconciling geometric and visual fidelity within a unified GT23D framework.

\textbf{(3) The limited retention of image priors underscores the inherent difficulty of transferring 2D knowledge into 3D generative spaces.}
Models such as DreamView show that leveraging strong pre-trained text-to-image priors can markedly enhance high-fidelity 3D synthesis under multi-view supervision, even surpassing empirical upper bounds in global quality (GQ) and detail richness (DR) (Tab.~\ref{tab:rankings}). However, models generating explicit 3D outputs (e.g., meshes or point clouds) fail to preserve similar benefits, despite initialization from image-pretrained backbones. We attribute this gap to \textit{semantic drift} during cross-modal transfer, where high-level visual and structural knowledge learned in 2D becomes attenuated in 3D adaptation. Overcoming this degradation remains an open challenge for achieving coherent and knowledge-consistent GT23D generation.

\textbf{(4) Current GT23D models remain predominantly perception-driven, prioritizing visual fidelity over genuine spatial reasoning and structural comprehension.}
While recent approaches have improved global text–3D alignment, they continue to struggle with accurately capturing spatial relations among multiple entities and representing hierarchical scene structures, as reflected in the relatively low scores in the \textit{Scene} category (Fig.~\ref{fig:baseline_radar}). This indicates that current approaches primarily rely on perceptual correlations rather than grounded spatial understanding. These observations highlight the necessity of advancing beyond pixel-level perceptual objectives toward incorporating structured spatial representations and relational reasoning, enabling GT23D models to internalize and reason about the underlying organization of 3D scenes rather than merely reproducing their appearances.

\textbf{(5) The tension between realism and imagination reflects an unresolved trade-off between fidelity and generalization.}
Existing methods face a persistent tension between generating photorealistic 3D content and producing creative or stylized results. Multi-view image–based models, such as the MVDream series, tend to generate stylized or cartoon-like outputs due to their 2D diffusion priors, whereas native 3D models achieve relatively higher realism yet still exhibit stylization. This phenomenon suggests that current 3D datasets inherently encode aesthetic biases. Building more diverse and photorealistic 3D datasets is therefore crucial to mitigating these biases and reconciling fidelity with generalization in future GT23D development.

\section{Conclusion}

We present \textbf{GT23D-Bench}, the first comprehensive benchmark tailored to the General Text-to-3D (GT23D) generation task, addressing long-standing challenges in dataset construction and evaluation standardization. Our benchmark delivers three key contributions that jointly advance the field: (1) a large-scale, well-structured 3D dataset comprising 400K high-quality shapes curated from Objaverse, enriched with multimodal annotations including 64-view renderings, depth and normal maps, and hierarchical textual descriptions; (2) a suite of holistic 3D-aware evaluation metrics covering 10 dimensions, designed to assess both \textit{Text-3D Alignment} and \textit{3D Visual Quality}, thereby capturing fine-grained aspects of semantics and geometry beyond conventional image-level assessments; and (3) actionable benchmarking insights from evaluating 8 state-of-the-art GT23D methods, providing a clear view of current capabilities and limitations.

Our extensive experiments demonstrate that GT23D-Bench exhibits strong correlation with human preference, highlighting its reliability and interpretability. Moreover, the benchmark reveals critical gaps in current models, particularly in multi-view consistency and shape completeness, which are often overlooked by existing evaluation tools. By filling these gaps, GT23D-Bench not only enables fair and rigorous comparison of GT23D methods but also sets a new standard for future research in this rapidly evolving area. We believe our benchmark will serve as a foundational resource to guide the development of more faithful, robust, and semantically aligned general text-to-3D generation.

\bibliographystyle{IEEEtran}
\bibliography{references}

\newpage
\appendices

\end{document}